\documentclass[letterpaper]{article} 
\usepackage[preprint]{aaai2027}  
\usepackage[hyphens]{url}  
\usepackage{graphicx} 
\usepackage{natbib}  
\usepackage{caption} 
\usepackage{algorithm}
\usepackage{algorithmicx}
\usepackage{algpseudocode}

\usepackage{newfloat}
\usepackage{listings}

\usepackage{amsfonts}
\usepackage{amsmath}
\usepackage{amssymb}
\usepackage{amsthm}
\usepackage{enumitem}
\usepackage{nicefrac}
\usepackage{microtype}
\usepackage{xcolor}
\usepackage{tikz}
\usetikzlibrary{arrows.meta,positioning,calc}

\DeclareCaptionStyle{ruled}{labelfont=normalfont,labelsep=colon,strut=off} 
\floatstyle{ruled}
\newfloat{listing}{tb}{lst}{}
\floatname{listing}{Listing}

\usepackage{booktabs}

\title{Kilobyte Models: Neural Networks as a Seed and a Quantized Latent}
\author {
    Sahil Rajesh Dhayalkar
}
\affiliations{
    Arizona State University\\
    sdhayalk@asu.edu
}

\newcommand{\R}{\mathbb{R}}
\newcommand{\vz}{\mathbf{z}}
\newcommand{\vone}{\mathbf{1}}
\newcommand{\norm}[1]{\left\lVert #1 \right\rVert}

\begin{document}

\maketitle

\begin{abstract}
The cost of storing and transmitting a trained neural network scales with its parameter count, a bottleneck for over-the-air updates, on-device libraries, and other bandwidth-bound deployments. We study an extreme form of model compression in which the deployable artifact is not the weights but a short recipe for regenerating them. Building on Mapping Networks, which express a network's weights as a nonlinear function of a compact trainable latent and a fixed random basis, we observe that only the latent need be stored, because the basis and initialization center are reproducible from an integer seed. A model becomes a seed together with a quantized latent, whose size is set by the latent dimension and bit width rather than the parameter count. We formalize this artifact and introduce a seeded block-wise basis that scales to networks whose projection cannot be held in memory. In our experiments, a mapped model is as accurate as the same network quantized aggressively to a few bits per weight, while taking far fewer bytes to store. Reaching the most aggressive bit widths depends on fine-tuning the latent with quantization in the loop. The results do not depend on the particular random basis, and a structured basis lets the weights be regenerated almost for free even for large networks.
\end{abstract}


\section{Introduction}
\label{sec:intro}

Storing and transmitting neural networks scales with parameter count, creating bottlenecks for bandwidth-constrained deployments like edge devices, over-the-air updates, and massively scaled adapters. In these scenarios, the binding constraint is storage size rather than inference cost. Standard compression techniques, such as pruning~\cite{han2015learning,frankle2019lottery}, quantization~\cite{jacob2018quantization,han2016deep}, low-rank factorization, and distillation~\cite{hinton2015distilling}, address this by shrinking the description of the original weight tensors. 

Alternatively, one can store a compact recipe to regenerate weights. Neural networks have low intrinsic dimensionality~\cite{denil2013predicting}, allowing optimization within random, low-dimensional subspaces~\cite{li2018intrinsic}. If a random seed generates this subspace, the model is entirely defined by that seed and a small coordinate vector, reducing the byte count. 

Building on Mapping Networks~\cite{sen2026mapping}, which represent weights as a nonlinear function of a small trainable latent vector and a fixed random basis, we propose an extreme compression paradigm. Since the random basis and initialization are deterministically regenerated from an integer seed, only the latent vector requires storage. Quantizing this latent to a few bits shrinks the entire artifact to a few kilobytes. We formalize this deployable artifact as a \emph{Kilobyte model} and evaluate its accuracy at extreme compression rates.

Our specific contributions are as follows:
\begin{itemize}[itemsep=0pt,topsep=0pt]
  \item We formalize the Kilobyte model artifact (seed, quantized latent, and normalization parameters), which is sized by latent dimension $d$ and bit width $b$ rather than the full network parameter count $P$ (Section~\ref{sec:artifact}).
  \item We introduce a scalable, seeded, block-wise random basis that avoids materializing dense projection matrices, regenerating weights efficiently across platforms (Section~\ref{sec:basis}).
  \item We extend this parameterization to \emph{fine-tuning}, using the latent as a compressed task adapter over a shared pretrained base network to produce a small-sized delta (Section~\ref{sec:finetune}).
  \item We demonstrate that Kilobyte models match aggressively quantized networks at a fraction of their size. For instance, a 4-bit model achieves $98.6\%$ on MNIST in just 2 KB, yielding a $26\times$ reduction over the quantized baseline 
  (Section~\ref{sec:cnn-results}).
\end{itemize}

\section{Related Work}
\label{sec:related}

\paragraph{Random subspaces and intrinsic dimension.}
Neural networks can be successfully optimized within random, low-dimensional affine subspaces, revealing a surprisingly small intrinsic dimension for most objective landscapes~\cite{li2018intrinsic,aghajanyan2020intrinsic,sen2026mapping}. The manifold hypothesis suggests that trained weights concentrate near a low-dimensional set, possessing robust empirical~\cite{denil2013predicting,sen2026mapping} and theoretical~\cite{fefferman2013testingmanifoldhypothesis} backing. This intrinsic dimension remains manageable even as models scale to billions of parameters~\cite{aghajanyan2020intrinsic}. While prior research used dense random subspaces primarily as analytical measurement tools, we repurpose them into a highly compact storage container. We achieve this by making the random projection reproducible from an integer seed.

\paragraph{Reparameterization-based compression.}
Our method relates to techniques storing models as a seed alongside a small set of trainable coefficients. Random projections have a long history in machine learning~\cite{rahimi2007random}, and recent methods apply them directly to network parameters. NOLA~\cite{koohpayegani2024nola} expresses weights as linear combinations of pseudo-random basis networks, while MCNC~\cite{thrash2025mcnc} constrains parameters to frozen nonlinear manifolds. Our parameterization differs critically and adapts from~\cite{sen2026mapping}. It is strictly nonlinear, centers the map at the standard initialization to encode a structural displacement rather than an absolute location, and utilizes latent quantization as the primary compression lever.

\paragraph{Weight-space compression.}
Traditional compression explicitly targets trained weights. Early methods~\cite{lecun1989optimal} led to unstructured pruning that removes weights by magnitude~\cite{han2015learning} or by identifying winning lottery tickets~\cite{frankle2019lottery,chen2020lottery}. Structured pruning offers hardware-friendly alternatives~\cite{wen2016learning}. Quantization reduces numerical precision through post-training methods~\cite{nagel2020up,yao2021hawq} or straight-through quantization-aware training~\cite{jacob2018quantization,bengio2013estimating,courbariaux2015binaryconnect}. Other techniques include weight clustering~\cite{han2016deep}, hashing~\cite{chen2015hashednets}, and knowledge distillation~\cite{hinton2015distilling,gou2021knowledge}. Our reparameterization paradigm is fundamentally complementary because it never represents the full weight set during optimization. Instead, standard compression techniques like quantization are shifted entirely to the compact latent space to amplify the overall size reduction.

\paragraph{Parameter-efficient adaptation and hypernetworks.}
Parameter-efficient fine-tuning adapts large models with minimal storage overhead. Standard adapters insert small bottleneck layers~\cite{houlsby2019parameter}, while prompt tuning optimizes hidden states~\cite{li2021prefix}. Methods like LoRA~\cite{hu2022lora}, QLoRA~\cite{dettmers2023qlora}, and GaLore~\cite{zhao2024galore} constrain updates or gradients to low-rank subspaces. VeRA~\cite{kopiczko2024vera} freezes random low-rank matrices and trains only scaling vectors. Hypernetworks~\cite{ha2017hypernetworks,chauhan2023brief} generate weights using a secondary network but do not prioritize minimal storage footprints. While our framework focuses on from-scratch training for extreme compression, it naturally accommodates adaptation. Centering the mapping on pretrained weights turns our latent vector into a kilobyte-sized adapter (Section~\ref{sec:finetune}), offering a hyper-compressed counterpart.

\section{Preliminaries: Mapping Networks}
\label{sec:prelim}

We follow the Mapping Network formulation introduced in~\cite{sen2026mapping}. Let $f_{\theta}$ be a target network with flattened trainable parameters $\theta \in \R^{P}$, where $P$ is the total number of such parameters, and write $f_{\theta}(x)$ for its output on input $x$. Ordinarily $\theta$ is optimized directly by minimizing a task loss $\mathcal{L}$ over a dataset. A Mapping Network instead introduces a trainable latent vector $\vz \in \R^{d}$ with $d \ll P$ and a fixed, differentiable map $g : \R^{d} \to \R^{P}$, and sets $\theta = g(\vz)$. Concretely,
\begin{equation}
\label{eq:map}
g(\vz) \;=\; \sigma\!\left( W_{0}\, \vz \;+\; \alpha \norm{\vz}_{2}^{2}\, \vone \;+\; b_{0} \right),
\end{equation}
where $W_{0} \in \R^{P \times d}$ is a fixed projection with near-orthonormal columns, $b_{0} \in \R^{P}$ is a fixed center, $\alpha \in \R$ is a small scalar modulation coefficient, $\vone \in \R^{P}$ is the all-ones vector, $\sigma$ is an element-wise nonlinearity taken to be $\tanh$, and $\norm{\cdot}_{2}$ is the Euclidean norm. The center $b_{0}$ is set to the target network's standard initialization, so that at $\vz = 0$ the generated parameters equal a well-scaled initial network and the latent learns a displacement from it. The term $\alpha \norm{\vz}_{2}^{2}$ is a scalar, added to every coordinate, that follows from the additive weight modulation of the original formulation and is negligible for the small $\alpha$ we use.

Training minimizes the task loss with respect to the latent only~\cite{sen2026mapping},
\begin{equation}
\label{eq:train}
\vz^{\star} \;=\; \arg\min_{\vz \in \R^{d}} \; \mathbb{E}_{(x,y)}\, \mathcal{L}\!\left( f_{g(\vz)}(x),\, y \right),
\end{equation}
with $W_{0}$, $b_{0}$, and $\alpha$ held fixed, so that gradients flow through $g$ into $\vz$ but never into $W_{0}$ or $b_{0}$. The target network is used only for its forward pass. The Mapping Theorem of~\cite{sen2026mapping} shows, under Lipschitz and smoothness assumptions on the loss and the map, that for a target parameter $\theta^{\star}$ on a low-dimensional manifold there exists a latent $\vz^{\star}$ with $g(\vz^{\star})$ arbitrarily close to $\theta^{\star}$, which is the existence statement that motivates optimizing in $\vz$.

\paragraph{Two training regimes.} ~\cite{sen2026mapping} distinguish two ways of applying the Mapping Network to a target, which they call \emph{Single Latent Vector Training} (SLVT) and \emph{Layer-Wise Training} (LWT). In SLVT, a single latent vector generates all of the target's mapped parameters at once, exactly as in Equations~\eqref{eq:map} and \eqref{eq:train}. In LWT the target is instead handled one layer at a time, where a layer is a single trainable module of the network's architecture (an individual convolutional or fully-connected layer): the mapped parameters are partitioned by layer, $\theta_{\mathrm{map}} = (\theta^{1}, \dots, \theta^{L})$, with $\theta^{\ell}$ the parameters of the $\ell$-th layer, $P_{\ell}$ their count, and $\sum_{\ell=1}^{L} P_{\ell} = P$, and each layer carries its own latent $\vz_{\ell} \in \R^{d_{\ell}}$, frozen projection $W_{0}^{\ell} \in \R^{P_{\ell} \times d_{\ell}}$, and center $b_{0}^{\ell}$:
\begin{equation}
\label{eq:map-lwt}
\theta^{\ell} \;=\; \sigma\!\left( W_{0}^{\ell}\, \vz_{\ell} \;+\; \alpha \norm{\vz_{\ell}}_{2}^{2}\, \vone \;+\; b_{0}^{\ell} \right), \qquad \ell = 1, \dots, L.
\end{equation}
SLVT is the special case where $L = 1$. LWT spends a larger total latent budget $D = \sum_{\ell} d_{\ell}$ in exchange for finer control, since the per-layer dimensions $d_{\ell}$ can be allocated in proportion to each layer's size or difficulty rather than sharing one latent across the whole network. Everything that follows, including the deployable artifact, the seeded basis, and latent quantization, applies unchanged to each per-layer block. To keep the notation light, we state it for a single latent and flag the layer-wise sum where it matters. ~\cite{sen2026mapping} also applies the same map beyond from-scratch training, to fine-tune a pretrained network by additive modulation of its weights, and we build on this in Section~\ref{sec:finetune}.

The property we exploit is structural rather than about accuracy. In Eq.~\eqref{eq:map}, the objects that carry learned information and the objects that are fixed are cleanly separated: only $\vz$ is trained, while $W_{0}$, $b_{0}$, and $\alpha$ are fixed the moment they are chosen. If those fixed objects can be regenerated on demand, the trained model is fully described by $\vz$.

\section{Kilobyte Models}
\label{sec:method}

\begin{figure}[t]
  \centering
  \begin{tikzpicture}[
      font=\small,
      box/.style={draw, rounded corners=2pt, minimum height=8mm, inner sep=4pt, align=center},
      seedbox/.style={box, fill=blue!6},
      lat/.style={box, fill=green!8},
      op/.style={box, fill=gray!8},
      >={Latex[length=2mm]}]
      
    \node[seedbox] (seed) {seed $s$\\(8 bytes)};
    \node[lat, right=6mm of seed] (z) {quantized latent\\$Q_{b}(\vz^{\star})$\ ($db/8$ bytes)};
    
    \path (seed.north) -- (z.north) coordinate[midway] (midN);
    \node[above=1mm of midN, blue!55!black, font=\scriptsize] {stored / transmitted artifact};
    \draw[thick, blue!45, rounded corners] ($(seed.north west)+(-2mm, 6mm)$) rectangle ($(z.south east)+(2mm, -2mm)$);
    
    \node[op, below=6mm of seed] (gen) {regenerate\\$W_{0}=\rho(s),\ b_{0}=\rho'(s)$};
    
    \node[op] (map) at ($(midN) + (0, -34mm)$) {$\theta=\sigma(W_{0}\vz+\alpha\norm{\vz}^2\vone+b_{0})$};
    
    \node[box, fill=orange!8, below=5mm of map] (net) {target\\$f_{\theta}(x)$};
    
    \draw[->] (seed) -- (gen);
    
    \path (map.north) ++(0, 8mm) coordinate (mapTop);
    \draw[->] (gen.south) -- (gen.south |- map.north);
\draw[->] (z.south) -- (z.south |- map.north);
    
    \draw[->] (map) -- (net);
    
  \end{tikzpicture}
  \caption{The Kilobyte model artifact. Only the seed $s$ and the quantized latent $Q_{b}(\vz^{\star})$ are stored or transmitted (boxed). On the receiving device, the seed regenerates the frozen projection $W_{0}$ and center $b_{0}$, the map of Eq.~\eqref{eq:map} produces the full parameter vector $\theta$, and the target network runs a standard forward pass. The artifact size is set by the latent dimension $d$ and bit width $b$, not by the parameter count $P$.}
  \label{fig:pipeline}
\end{figure}
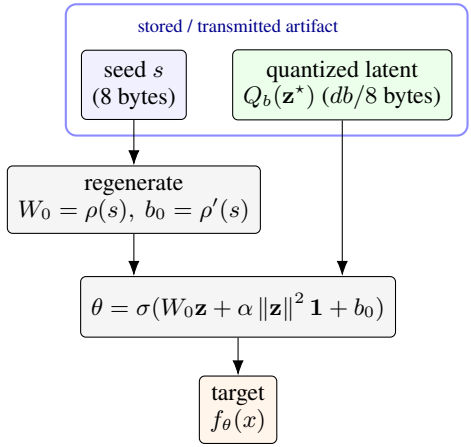

\subsection{The deployable artifact}
\label{sec:artifact}

The parameters of a real network include normalization affine parameters (the scale and shift of group, layer, or batch normalization), whose standard initialization is the all-ones and all-zeros vectors. These lie on the boundary of the range of $\tanh$ and cannot be produced by Eq.~\eqref{eq:map}. Furthermore, they are also few in number. We therefore partition the target parameters as $\theta = (\theta_{\mathrm{map}}, \theta_{\mathrm{norm}})$, where $\theta_{\mathrm{map}} \in \R^{P}$ are the convolutional and linear weights and biases, which form the overwhelming majority and are generated by the map, and $\theta_{\mathrm{norm}} \in \R^{P_{n}}$ are the normalization parameters, with $P_{n} \ll P$, which are trained directly and stored verbatim.

A trained Kilobyte model is then the triple
\begin{equation}
\label{eq:artifact}
\mathcal{A} \;=\; \bigl(\, s,\; Q_{b}(\vz^{\star}),\; \theta_{\mathrm{norm}} \,\bigr),
\end{equation}
where $s$ is the integer seed that generates $W_{0}$ and $b_{0}$, $Q_{b}$ is a $b$-bit quantizer, and $\vz^{\star}$ is the trained latent. To deploy the model, a device regenerates $W_{0}$ and $b_{0}$ from $s$, dequantizes the latent, evaluates Eq.~\eqref{eq:map} to obtain $\theta_{\mathrm{map}}$, reattaches $\theta_{\mathrm{norm}}$, and runs the ordinary forward pass of $f_{\theta}$ (Figure~\ref{fig:pipeline}). The size of the artifact in bytes is
\begin{equation}
\label{eq:bytes}
|\mathcal{A}| \;=\; \underbrace{\frac{d\, b}{8}}_{\text{latent payload}} \;+\; \underbrace{4}_{\text{scale}} \;+\; \underbrace{8}_{\text{seed}} \;+\; \underbrace{4 P_{n}}_{\text{norm params}},
\end{equation}
where the dominant term for the budgets we consider is the latent payload $d b / 8$. 
The remaining terms are stored in full precision: each layer's quantization scale is a single $32$-bit float, giving $4L$ bytes across the $L$ layers (and just $4$ bytes under SLVT, where $L = 1$); each of the $P_{n}$ directly-trained normalization parameters is a $32$-bit float ($4$ bytes, hence $4 P_{n}$); and the seed is a single $64$-bit integer ($8$ bytes), regardless of the number of layers, since one master seed generates every $W_{0}^{\ell}$.
Inference is unchanged relative to the target network, because after regeneration, the weights are dense floating-point values (the compression is of the stored and transmitted description, not of the compute).

\subsection{Seeded basis generation at scale}
\label{sec:basis}

For the artifact to remain small, the receiver must feasibly regenerate $W_{0} \in \R^{P \times d}$ from the seed alone, even when $W_{0}$ exceeds commodity accelerator memory (e.g., tens of gigabytes for a 1.2M parameter target with $d = 16{,}384$). Unlike Mapping Networks~\cite{sen2026mapping}, we use a reproducible, near-orthonormal projection generated in row blocks to avoid full matrix materialization.

Writing $W_{0} = [B_{1}^{\top}, \dots, B_{K}^{\top}]^{\top}$ as a stack of $K$ row blocks, each block $B_{k}$ is drawn from a pseudo-random generator seeded by a mixing function of the seed $s$ and block index $k$. Entries are either Gaussian, $[B_{k}]_{ij} \sim \mathcal{N}(0, 1/P)$, or Rademacher, $[B_{k}]_{ij} \in \{-1/\sqrt{P}, +1/\sqrt{P}\}$. These i.i.d. entries allow independent block regeneration from $(s, k)$, avoiding the full materialization required by exact orthogonalization. The $1/\sqrt{P}$ scaling ensures expected near-orthonormality, $\mathbb{E}[W_{0}^{\top} W_{0}] = I_{d}$, meaning the map $\vz \mapsto W_{0}\vz$ approximately preserves inner products for large $P$.

This deterministic generation guarantees cross-device portability. For LWT, a single master seed incorporates the layer index $\ell$ to generate all $L$ projections $W_{0}^{\ell}$. Memory permitting, $W_{0}$ can be cached, streamed, or regenerated dynamically using a custom block-wise differentiable operator to compute $\vz \mapsto W_{0}\vz$ and its transpose.

\subsection{Latent quantization}
\label{sec:quant}

The latent payload shrinks directly with the bit width $b$, so we store $\vz$ with standard symmetric uniform quantization~\cite{jacob2018quantization} and consider two variants. Post-training quantization (PTQ) simply rounds the already-trained latent to $b$ bits. Quantization-aware training (QAT) instead fine-tunes the latent with the rounding simulated in the loop, so its values settle where they survive quantization as gradients are passed through the rounding with a straight-through estimator~\cite{bengio2013estimating}. Because the latent has only $d$ entries, this is inexpensive and touches a single small vector rather than the full weight set. Algorithms for encoding and decoding are provided in the Appendix.

\subsection{Fine-tuning as a kilobyte delta}
\label{sec:finetune}

While the previous sections describe training from scratch, our framework naturally extends to fine-tuning, similar to~\cite{sen2026mapping}. By replacing the initialization center $b_{0}$ in Eq.~\eqref{eq:map} with a frozen pretrained parameter vector $\theta_{\mathrm{pre}}$ and using an identity activation, the map generates a displacement from the pretrained solution:
\begin{equation}
\label{eq:finetune}
\theta_{\mathrm{ft}} \;=\; \theta_{\mathrm{pre}} \;+\; W_{0}\, \vz,
\end{equation}
Here, only $\vz$ is trained. The latent encodes a task-specific \emph{adapter} within the frozen random subspace, similar to LoRA~\cite{hu2022lora} and VeRA~\cite{kopiczko2024vera}, but reduced strictly to a seed and a quantized latent. The seeded basis generation, quantization strategies, and training regimes apply unchanged.

The byte accounting, however, must adapt. Because $\theta_{\mathrm{pre}}$ is a specific pretrained model rather than a seed-regenerable initialization, it is treated as a shared prerequisite assumed to be already present on the deployment device. A fine-tuned Kilobyte model is therefore defined as:
\begin{equation}
\label{eq:artifact-ft}
\mathcal{A}_{\mathrm{ft}} \;=\; \bigl(\, h,\; s,\; \{Q_{b}(\vz_{\ell}^{\star})\}_{\ell=1}^{L},\; \theta_{\mathrm{norm}} \,\bigr),
\end{equation}
where $h$ is a short identifier (hash) of the base model. This hash binds the delta to its training base, ensuring it is applied to the correct pretrained weights.

\section{Experiments}
\label{sec:exp}

We evaluate all targets under both SLVT and LWT regimes (Section~\ref{sec:prelim}) using identical byte accounting. We report results for both from-scratch targets and a fine-tuning adapter (Section~\ref{sec:finetune}) evaluated on a pretrained ResNet-50. Mapped and baseline models are trained via Adam~\cite{Kingma2014AdamAM}. Experiments are implemented in PyTorch with CUDA on an NVIDIA RTX 4060 GPU across 3 random seeds. We report mean metrics; corresponding standard deviations and hyperparameters are detailed in the Appendix.

\subsection{Seeded basis is reproducible and near-orthonormal}
\label{sec:exp-basis}

Before evaluating accuracy we verify that the seeded basis of Section~\ref{sec:basis} behaves as the parameterization requires. Table~\ref{tab:basis} summarizes the Gram matrix $W_{0}^{\top}W_{0}$ across the parameter counts of our targets: the columns have unit norm to within $0.1\%$, and the off-diagonal correlations are small and shrink as $1/\sqrt{P}$, so larger targets are closer to orthonormal. The Gaussian and Rademacher constructions agree at every $P$, so this does not depend on the particular random basis. 
We also verify the generator. The cached and regenerated back-ends produce identical projections, a fixed seed reproduces the basis exactly, and a device-independent generator is bit-identical across runs, in all three cases to a maximum difference of $0$. The block-wise operator agrees with the closed-form transpose to $2.6 \times 10^{-6}$. A seed is therefore a portable description of the model.
For why the projection is taken to be orthogonal in the first place, please refer to~\cite{sen2026mapping}.

\begin{table}[t]
  \caption{Near-orthonormality of the seeded basis: Gram statistics of $W_{0}^{\top}W_{0}$ over a sample of $256$ columns, for the Gaussian and Rademacher constructions at three parameter counts $P$.}
  \label{tab:basis}
  \centering
  \small
  \setlength{\tabcolsep}{5pt} 
  \begin{tabular}{lcccc}
    \toprule
    Construction & P & Mean col- & Mean           & Max            \\
                 &  & umn norm & $|$off-diag$|$ & $|$off-diag$|$ \\
    \midrule
    Gaussian &$105{,}866$ & $0.99983$ & $0.00245$ & $0.01242$ \\
    Gaussian &$538{,}081$ & $0.99991$ & $0.00108$ & $0.00571$ \\
    Gaussian &$5{,}044{,}942$ & $0.99998$ & $0.00035$ & $0.00177$ \\
    \midrule
    Rademacher &$105{,}866$ & $0.99999$ & $0.00244$ & $0.01319$ \\
    Rademacher &$538{,}081$ & $1.00010$ & $0.00109$ & $0.00595$ \\
    Rademacher &$5{,}044{,}942$ & $0.99905$ & $0.00035$ & $0.00180$ \\
    \bottomrule
  \end{tabular}
\end{table}

\subsection{CNN Results}
\label{sec:cnn-results}

\begin{figure}[t]
  \centering
  \includegraphics[width=\linewidth]{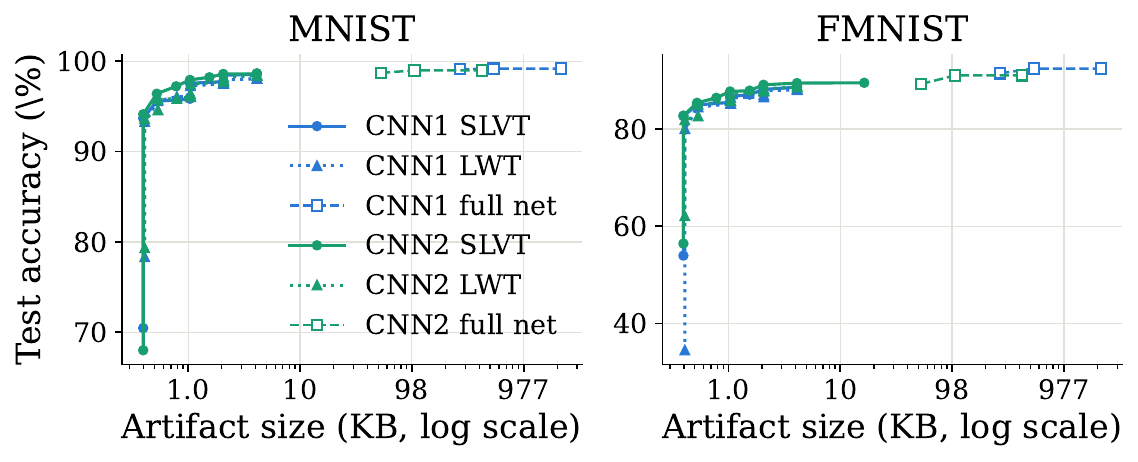}
  \caption{Accuracy against artifact size for the CNN targets.}
  \label{fig:cnn}
\end{figure}

We evaluate the Kilobyte model artifact in the same classification setting of~\cite{sen2026mapping}: two convolutional targets, CNN1 ($538{,}081$ parameters) and CNN2 ($105{,}866$), on MNIST~\cite{lecun1998mnist} and FashionMNIST~\cite{fashionmnist}. Both are plain convolutional networks with no normalization layers, so the artifact stores no separate normalization parameters. We sweep the latent budget $d \in \{1024, 2048, 4096\}$ at full precision and at eight and four bits under both PTQ and QAT, and both training regimes: SLVT, and LWT with $d$ split across the four conv/linear layers in proportion to their size (which leaves the two regimes essentially the same artifact size at a given $d$). Tables~\ref{tab:cnn} and~\ref{tab:cnn-bytes} and Figure~\ref{fig:cnn} report the results.

\begin{table}[h!]
  \caption{CNN Results: Test accuracy (\%) for the two targets trained conventionally (``full'') and as mapped Kilobyte models under SLVT and LWT, at latent budget $d$ and int8/int4 under PTQ and QAT, on MNIST and FashionMNIST.}
  \label{tab:cnn}
  \centering
  \small
  \setlength{\tabcolsep}{4pt}
  \begin{tabular}{l r c c c c c}
    \toprule
    & & \multicolumn{5}{c}{Test accuracy (\%)} \\
    \cmidrule(lr){3-7}
    & & fp32 & \multicolumn{2}{c}{int8} & \multicolumn{2}{c}{int4} \\
    \cmidrule(lr){4-5}\cmidrule(lr){6-7}
    Model & $d$ & & PTQ & QAT & PTQ & QAT \\
    \midrule
    \multicolumn{7}{l}{\emph{MNIST}} \\
    \shortstack[l]{CNN1 full \\ $P=538{,}081$} & --- & $99.20$ & $99.18$ & --- & $99.12$ & --- \\
    CNN1 SLVT & $1024$ & $95.97$ & $95.85$ & $96.00$ & $87.12$ & $95.56$ \\
    CNN1 SLVT & $2048$ & $97.73$ & $97.74$ & $97.80$ & $87.16$ & $97.57$ \\
    CNN1 SLVT & $4096$ & $98.51$ & $98.57$ & $98.61$ & $96.20$ & $98.45$ \\
    CNN1 LWT  & $1024$ & $96.09$ & $96.11$ & $96.00$ & $92.41$ & $95.61$ \\
    CNN1 LWT  & $2048$ & $97.56$ & $97.56$ & $97.53$ & $95.65$ & $97.26$ \\
    CNN1 LWT  & $4096$ & $98.06$ & $98.07$ & $98.08$ & $97.09$ & $97.94$ \\
    \shortstack[l]{CNN2 full \\ $P=105{,}866$} & --- & $98.99$ & $98.99$ & --- & $98.72$ & --- \\
    CNN2 SLVT & $1024$ & $96.73$ & $96.73$ & $96.68$ & $91.09$ & $96.41$ \\
    CNN2 SLVT & $2048$ & $98.10$ & $98.13$ & $98.08$ & $94.46$ & $97.94$ \\
    CNN2 SLVT & $4096$ & $98.66$ & $98.67$ & $98.61$ & $97.49$ & $\mathbf{98.60}$ \\
    CNN2 LWT  & $1024$ & $96.49$ & $96.38$ & $96.37$ & $94.60$ & $95.79$ \\
    CNN2 LWT  & $2048$ & $97.81$ & $97.80$ & $97.72$ & $96.05$ & $97.46$ \\
    CNN2 LWT  & $4096$ & $98.46$ & $98.47$ & $98.42$ & $96.83$ & $98.22$ \\
    \midrule
    \multicolumn{7}{l}{\emph{FashionMNIST}} \\
    \shortstack[l]{CNN1 full \\ $P=538{,}081$} & --- & $92.45$ & $92.46$ & --- & $91.47$ & --- \\
    CNN1 SLVT & $1024$ & $85.55$ & $85.50$ & $85.49$ & $78.16$ & $84.97$ \\
    CNN1 SLVT & $2048$ & $86.86$ & $86.90$ & $86.81$ & $78.27$ & $86.67$ \\
    CNN1 SLVT & $4096$ & $88.70$ & $88.62$ & $88.83$ & $85.05$ & $88.16$ \\
    CNN1 LWT  & $1024$ & $85.34$ & $85.21$ & $85.29$ & $78.26$ & $84.51$ \\
    CNN1 LWT  & $2048$ & $86.76$ & $86.55$ & $86.97$ & $79.31$ & $86.49$ \\
    CNN1 LWT  & $4096$ & $88.13$ & $88.08$ & $88.27$ & $81.36$ & $87.90$ \\
    \shortstack[l]{CNN2 full \\ $P=105{,}866$} & --- & $91.06$ & $91.02$ & --- & $89.28$ & --- \\
    CNN2 SLVT & $1024$ & $85.88$ & $85.81$ & $85.96$ & $74.16$ & $85.40$ \\
    CNN2 SLVT & $2048$ & $87.80$ & $87.85$ & $87.84$ & $85.13$ & $87.74$ \\
    CNN2 SLVT & $4096$ & $89.52$ & $89.46$ & $89.46$ & $87.84$ & $\mathbf{89.11}$ \\
    CNN2 LWT  & $1024$ & $85.88$ & $85.74$ & $85.89$ & $82.67$ & $85.06$ \\
    CNN2 LWT  & $2048$ & $87.72$ & $87.75$ & $87.92$ & $83.75$ & $87.19$ \\
    CNN2 LWT  & $4096$ & $88.67$ & $88.61$ & $88.81$ & $87.15$ & $88.32$ \\
    \bottomrule
  \end{tabular}
\end{table}

\begin{table}[h!]
  \caption{CNN artifact sizes (bytes). Mapped sizes depend only on $d$ and are identical across CNN1/CNN2 and both datasets, and LWT adds $12$ bytes (one extra scale per layer).}
  \label{tab:cnn-bytes}
  \centering
  \small
  \begin{tabular}{l r r r}
    \toprule
    Configuration & fp32 & int8 & int4 \\
    \midrule
    SLVT, $d=1024$ & $4{,}108$ & $1{,}036$ & $524$ \\
    SLVT, $d=2048$ & $8{,}204$ & $2{,}060$ & $1{,}036$ \\
    SLVT, $d=4096$ & $16{,}396$ & $4{,}108$ & $2{,}060$ \\
    CNN1 full & $2{,}152{,}324$ & $538{,}113$ & $269{,}072$ \\
    CNN2 full & $423{,}464$ & $105{,}898$ & $52{,}965$ \\
    \bottomrule
  \end{tabular}
\end{table}

The results show several key patterns. First, QAT performs much better than PTQ at low bit widths. For example, CNN1 on MNIST at $d=4096$ improves from 96.2\% (PTQ) to 98.5\% (QAT) at four bits. Second, eight-bit quantization loses almost no accuracy compared to full precision. Third, mapped models achieve high accuracy with a tiny footprint. CNN2 reaches 98.60\% on MNIST using just 2,060 bytes at four-bit QAT. Fourth, mapped models trail full networks slightly on FashionMNIST due to latent capacity limits, but larger latent dimensions help close this gap. Finally, SLVT and LWT methods perform similarly. SLVT has a small advantage at larger latent sizes and holds the best four-bit QAT scores. However, LWT is better for low-precision PTQ because it scales each layer separately. This makes LWT the better option when fine-tuning is not possible.

\subsection{LSTM Results}
\label{sec:lstm-results}

\begin{figure}[t]
  \centering
  \includegraphics[width=0.75\linewidth]{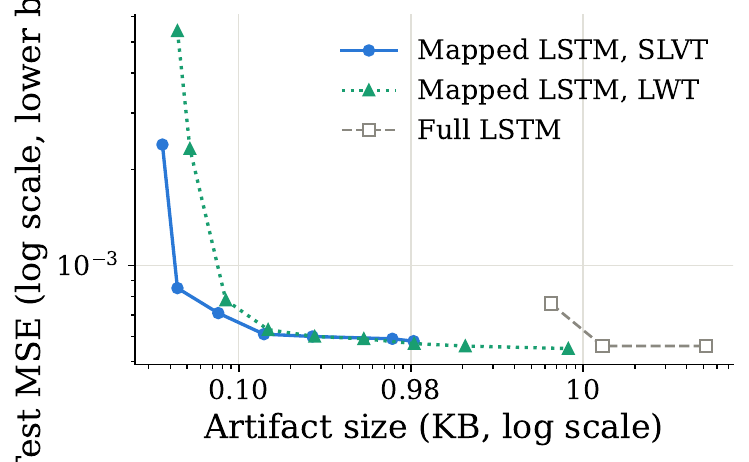}
  \caption{Test MSE against artifact size on the air-quality task (both axes log, lower MSE is better)}
  \label{fig:lstm}
\end{figure}

We repeat the study on a recurrent target and a regression task, following the time-series experiment of~\cite{sen2026mapping}. The target is a single-layer LSTM with $12{,}949$ parameters trained on the Beijing air-quality dataset~\cite{liang2015pm25}. The metric is test Mean Squared Error (MSE), where lower is better. As before, we sweep the latent dimension $d \in \{256, 1024, 2048\}$ and the bit width under both quantizers, against the conventionally trained LSTM compressed by post-training weight quantization. Tables~\ref{tab:lstm} and~\ref{tab:lstm-bytes} and Figure~\ref{fig:lstm} report the results.

The regression results match the classification patterns. Mapped models achieve similar error rates to the full LSTM but at a much smaller size. For instance, a four-bit QAT model at $d=256$ gets an MSE of 0.00061 using only 140 bytes. This is about 370 times smaller than the full-precision LSTM. As seen before, QAT is much better than PTQ for lower bit widths, while eight-bit quantization causes no accuracy loss.


\begin{table}[h!]
  \caption{LSTM Results on the Beijing air-quality dataset: test MSE ($\times10^{-3}$) for the LSTM trained conventionally (``full'') and as a mapped Kilobyte model under SLVT and LWT, at latent budget $d$ and int8/int4 under PTQ and QAT.}
  \label{tab:lstm}
  \centering
  \small
  \setlength{\tabcolsep}{5pt}
  \begin{tabular}{l r c c c c c}
    \toprule
    & & \multicolumn{5}{c}{Test MSE ($\times10^{-3}$, lower is better)} \\
    \cmidrule(lr){3-7}
    & & fp32 & \multicolumn{2}{c}{int8} & \multicolumn{2}{c}{int4} \\
    \cmidrule(lr){4-5}\cmidrule(lr){6-7}
    Model & $d$ & & PTQ & QAT & PTQ & QAT \\
    \midrule
    \shortstack[l]{LSTM full \\ $P=12{,}949$} & --- & 0.56 & 0.56 & --- & 0.76 & --- \\
    LSTM SLVT & $256$  & 0.60 & 0.60 & 0.60 & 1.26 & 0.61 \\
    LSTM SLVT & $1024$ & 0.58 & 0.58 & 0.59 & 0.90 & 0.64 \\
    LSTM SLVT & $2048$ & 0.59 & 0.58 & 0.62 & 0.81 & 0.61 \\
    LSTM LWT  & $256$  & 0.59 & 0.61 & 0.60 & 1.12 & 0.63 \\
    LSTM LWT  & $1024$ & 0.56 & 0.57 & 0.59 & 0.64 & 0.59 \\
    LSTM LWT  & $2048$ & 0.55 & 0.56 & 0.58 & 0.62 & $\mathbf{0.58}$ \\
    \bottomrule
  \end{tabular}
\end{table}

\begin{table}[h!]
  \caption{LSTM artifact sizes (bytes). Mapped sizes depend only on $d$, and LWT adds $8$ bytes (one extra scale per layer).}
  \label{tab:lstm-bytes}
  \centering
  \small
  \begin{tabular}{l r r r}
    \toprule
    Configuration & fp32 & int8 & int4 \\
    \midrule
    SLVT, $d=256$ & $1{,}036$ & $268$ & $140$ \\
    SLVT, $d=1024$ & $4{,}108$ & $1{,}036$ & $524$ \\
    SLVT, $d=2048$ & $8{,}204$ & $2{,}060$ & $1{,}036$ \\
    LSTM full & $51{,}796$ & $12{,}973$ & $6{,}498$ \\
    \bottomrule
  \end{tabular}
\end{table}

\subsection{MLP Results}
\label{sec:mlp-results}
\begin{figure}[h!]
  \centering
  \includegraphics[width=\linewidth]{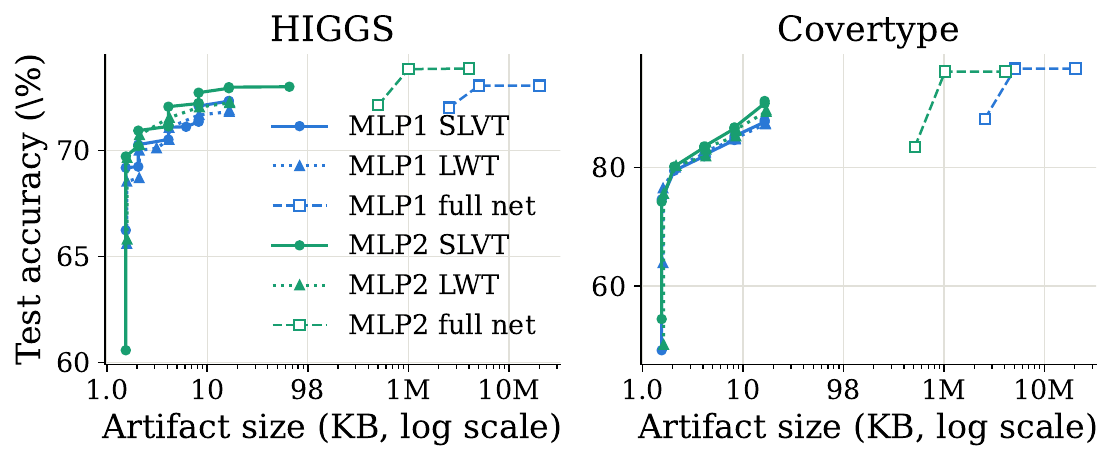}
  \caption{Accuracy against artifact size for the MLP targets.}
  \label{fig:mlp}
\end{figure}

We evaluate fully-connected targets on tabular data. We use two standard tabular benchmarks, the binary particle-physics dataset HIGGS~\cite{baldi2014higgs} (28 features, and we use a $600$k-row subset) and the seven-class forest Covertype dataset~\cite{blackard1999covertype} (54 features), and two three-hidden-layer MLPs: MLP1 with about five million parameters and MLP2 with about one million. Because a dense $P \times d$ projection for a five-million-parameter target is tens of gigabytes, here we generate the weights with the structured implicit basis of Section~\ref{sec:basis}. The sweep and the metric (test accuracy) are otherwise identical to the CNN study. Tables~\ref{tab:mlp} and~\ref{tab:mlp-bytes} and Figure~\ref{fig:mlp} report the results.

\begin{table}[h!]
  \caption{MLP Results on two tabular datasets: test accuracy (\%) for MLP1 and MLP2 trained conventionally (``full'') and as mapped Kilobyte models under SLVT and LWT, at latent budget $d$ and int8/int4 under PTQ and QAT.}
  \label{tab:mlp}
  \centering
  \small
  \setlength{\tabcolsep}{3.5pt}
  \begin{tabular}{l r c c c c c}
    \toprule
    & & \multicolumn{5}{c}{Test accuracy (\%)} \\
    \cmidrule(lr){3-7}
    & & fp32 & \multicolumn{2}{c}{int8} & \multicolumn{2}{c}{int4} \\
    \cmidrule(lr){4-5}\cmidrule(lr){6-7}
    Model & $d$ & & PTQ & QAT & PTQ & QAT \\
    \midrule
    \multicolumn{7}{l}{\emph{HIGGS} (binary)} \\
    \shortstack[l]{MLP1 full \\ $P=5{,}044{,}942$} & --- & $73.06$ & $73.06$ & --- & $72.02$ & --- \\
    MLP1 SLVT & $4096$ & $70.53$ & $70.52$ & $70.52$ & $69.24$ & $70.28$ \\
    MLP1 SLVT & $8192$ & $71.36$ & $71.36$ & $71.38$ & $68.86$ & $71.10$ \\
    MLP1 SLVT & $16384$ & $72.34$ & $72.34$ & $72.33$ & $70.85$ & $72.10$ \\
    MLP1 LWT  & $4096$ & $70.28$ & $70.33$ & $70.27$ & $68.71$ & $70.01$ \\
    MLP1 LWT  & $8192$ & $71.13$ & $71.14$ & $71.17$ & $70.51$ & $71.08$ \\
    MLP1 LWT  & $16384$ & $71.82$ & $71.83$ & $71.90$ & $69.77$ & $71.68$ \\
    \shortstack[l]{MLP2 full \\ $P=1{,}003{,}102$} & --- & $73.87$ & $73.84$ & --- & $72.14$ & --- \\
    MLP2 SLVT & $4096$ & $71.24$ & $71.20$ & $71.19$ & $70.21$ & $70.94$ \\
    MLP2 SLVT & $8192$ & $72.26$ & $72.22$ & $72.24$ & $71.14$ & $72.07$ \\
    MLP2 SLVT & $16384$ & $73.01$ & $72.99$ & $72.94$ & $70.45$ & $\mathbf{72.73}$ \\
    MLP2 LWT  & $4096$ & $70.95$ & $70.92$ & $70.95$ & $70.28$ & $70.74$ \\
    MLP2 LWT  & $8192$ & $71.70$ & $71.72$ & $71.70$ & $70.70$ & $71.55$ \\
    MLP2 LWT  & $16384$ & $72.23$ & $72.25$ & $72.29$ & $70.98$ & $72.05$ \\
    \midrule
    \multicolumn{7}{l}{\emph{Covertype} (7-class)} \\
    \shortstack[l]{MLP1 full \\ $P=5{,}093{,}927$} & --- & $96.65$ & $96.62$ & --- & $88.16$ & --- \\
    MLP1 SLVT & $4096$ & $82.12$ & $81.88$ & $82.10$ & $70.42$ & $79.43$ \\
    MLP1 SLVT & $8192$ & $84.71$ & $84.60$ & $84.82$ & $69.49$ & $82.02$ \\
    MLP1 SLVT & $16384$ & $87.76$ & $87.64$ & $87.94$ & $75.96$ & $85.25$ \\
    MLP1 LWT  & $4096$ & $82.68$ & $82.50$ & $82.71$ & $73.44$ & $79.99$ \\
    MLP1 LWT  & $8192$ & $84.85$ & $84.80$ & $84.90$ & $66.99$ & $82.37$ \\
    MLP1 LWT  & $16384$ & $87.31$ & $87.28$ & $87.39$ & $74.76$ & $84.73$ \\
    \shortstack[l]{MLP2 full \\ $P=1{,}024{,}807$} & --- & $96.13$ & $96.12$ & --- & $83.43$ & --- \\
    MLP2 SLVT & $4096$ & $83.43$ & $83.36$ & $83.53$ & $70.50$ & $80.13$ \\
    MLP2 SLVT & $8192$ & $86.62$ & $86.52$ & $86.73$ & $65.50$ & $83.37$ \\
    MLP2 SLVT & $16384$ & $91.04$ & $90.77$ & $91.18$ & $73.04$ & $\mathbf{86.64}$ \\
    MLP2 LWT  & $4096$ & $83.29$ & $83.03$ & $83.35$ & $71.42$ & $80.28$ \\
    MLP2 LWT  & $8192$ & $86.23$ & $86.16$ & $86.38$ & $74.83$ & $81.88$ \\
    MLP2 LWT  & $16384$ & $89.46$ & $89.38$ & $89.63$ & $72.75$ & $85.24$ \\
    \bottomrule
  \end{tabular}
\end{table}

The MLP results show consistent compression benefits. On the HIGGS dataset, mapped models closely match the full network's accuracy using only a fraction of the memory. On Covertype, the mapped models fall slightly behind the full network but improve quickly as the latent dimension increases. This shows that any accuracy gap is due to limited latent capacity. Once again, QAT is crucial for low bit widths, while eight-bit quantization incurs minimal performance drop. LWT and SLVT perform similarly on these tabular tasks, though SLVT keeps a slight advantage.

\begin{table}[h!]
  \caption{MLP artifact sizes (bytes). Mapped (SLVT) sizes depend only on $d$ and are identical across models and datasets.}
  \label{tab:mlp-bytes}
  \centering
  \small
  \setlength{\tabcolsep}{5pt}
  \begin{tabular}{l r r r}
    \toprule
    Configuration & fp32 & int8 & int4 \\
    \midrule
    SLVT, $d=4096$ & $16{,}396$ & $4{,}108$ & $2{,}060$ \\
    SLVT, $d=8192$ & $32{,}780$ & $8{,}204$ & $4{,}108$ \\
    SLVT, $d=16384$ & $65{,}548$ & $16{,}396$ & $8{,}204$ \\
    MLP1 full (HIGGS) & $20{,}179{,}768$ & $5{,}044{,}974$ & $2{,}522{,}503$ \\
    MLP1 full (Covertype) & $20{,}375{,}708$ & $5{,}093{,}959$ & $2{,}546{,}995$ \\
    MLP2 full (HIGGS) & $4{,}012{,}408$ & $1{,}003{,}134$ & $501{,}583$ \\
    MLP2 full (Covertype) & $4{,}099{,}228$ & $1{,}024{,}839$ & $512{,}435$ \\
    \bottomrule
  \end{tabular}
\end{table}

\subsection{Fine-tuning Results}
\label{sec:finetune-results}

\begin{figure}[t]
  \centering
  \includegraphics[width=0.75\linewidth]{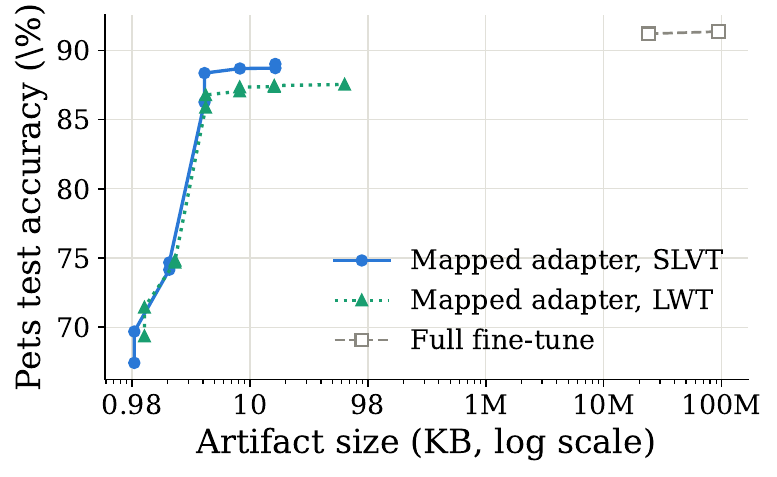}
  \caption{Pets accuracy against artifact size for the fine-tuned adapter, SLVT and LWT, against the full fine-tuned ResNet-50 compressed by weight quantization.}
  \label{fig:finetune}
\end{figure}

We now evaluate the fine-tuning regime of Section~\ref{sec:finetune}, where the Kilobyte model is a task adapter over a pretrained base. The base is an ImageNet-pretrained ResNet-50~\cite{he2016resnet} ($23{,}583{,}845$ parameters) with a fresh $37$-way head, adapted to the Oxford-IIIT Pet dataset~\cite{parkhi2012cats} ($3{,}680$ training and $3{,}669$ test images at $224{\times}224$) by training only the mapped delta of Eq.~\eqref{eq:finetune}. BatchNorm is frozen as part of the referenced base, so the artifact is just the seed and quantized latent, no verbatim weights. As for the MLP, the megaparameter target uses the structured basis. We sweep $d \in \{2048, 8192, 16384\}$ under both regimes and quantizers (fp32/int8/int4), training the latent with gradient clipping, which is needed for stability at the largest budget. The baseline is full fine-tuning of all weights, compressed post-training. Tables~\ref{tab:finetune} and~\ref{tab:finetune-bytes} and Figure~\ref{fig:finetune} report the results.

\begin{table}[t]
  \caption{Fine-tuning Results: Pets test accuracy (\%) for an ImageNet-pretrained ResNet-50 adapted by a mapped delta under SLVT and LWT, at latent budget $d$ and int8/int4 under PTQ and QAT. The baseline is full fine-tuning.}
  \label{tab:finetune}
  \centering
  \small
  \setlength{\tabcolsep}{4pt}
  \begin{tabular}{l r c c c c c}
    \toprule
    & & \multicolumn{5}{c}{Test accuracy (\%)} \\
    \cmidrule(lr){3-7}
    & & fp32 & \multicolumn{2}{c}{int8} & \multicolumn{2}{c}{int4} \\
    \cmidrule(lr){4-5}\cmidrule(lr){6-7}
    Method & $d$ & & PTQ & QAT & PTQ & QAT \\
    \midrule
    Full fine-tune & --- & $91.36$ & $91.20$ & --- & $2.94$ & --- \\
    \midrule
    SLVT & $2048$  & $74.30$ & $74.52$ & $74.03$ & $62.96$ & $69.66$ \\
    SLVT & $8192$  & $88.69$ & $88.69$ & $88.31$ & $86.26$ & $\mathbf{88.36}$ \\
    SLVT & $16384$ & $88.93$ & $89.02$ & $88.72$ & $88.14$ & $87.57$ \\
    LWT  & $2048$  & $70.99$ & $70.97$ & $70.70$ & $67.04$ & $67.26$ \\
    LWT  & $8192$  & $87.38$ & $87.35$ & $87.05$ & $85.88$ & $86.78$ \\
    LWT  & $16384$ & $87.54$ & $87.38$ & $87.46$ & $86.05$ & $86.32$ \\
    \bottomrule
  \end{tabular}
\end{table}

\begin{table}[t]
  \caption{Fine-tuning artifact sizes (bytes).}
  \label{tab:finetune-bytes}
  \centering
  \small
  \begin{tabular}{l r r r}
    \toprule
    Configuration & fp32 & int8 & int4 \\
    \midrule
    SLVT, $d=2048$ & $8{,}212$ & $2{,}068$ & $1{,}044$ \\
    SLVT, $d=8192$ & $32{,}788$ & $8{,}212$ & $4{,}116$ \\
    SLVT, $d=16384$ & $65{,}556$ & $16{,}404$ & $8{,}212$ \\
    Full fine-tune & $94{,}335{,}380$ & $23{,}956{,}065$ & $12{,}190{,}721$ \\
    \bottomrule
  \end{tabular}
\end{table}

The adapter acts as a highly effective fine-tuning tool. It comes within three accuracy points of full fine-tuning while being thousands of times smaller. For example, a four-bit QAT adapter uses just 4,116 bytes to reach 88.4\% accuracy, compared to a 94-megabyte full network. Unlike the full ResNet-50, which drops to 2.9\% accuracy under four-bit PTQ, the mapped adapter easily survives aggressive quantization. In this setting, SLVT performs slightly better than LWT. LWT allocates capacity based on layer size, which starves the tiny but crucial task head. SLVT avoids this by letting the head use the entire shared budget. 

\begin{figure}[h!]
  \centering
  \includegraphics[width=0.75\linewidth]{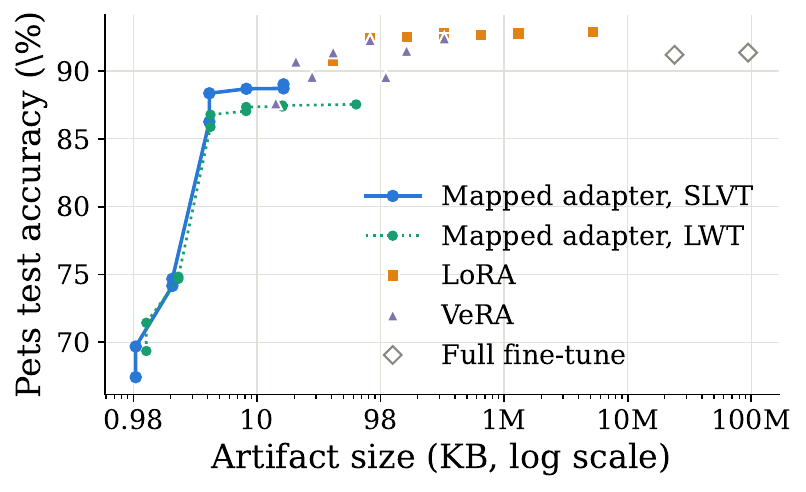}
  \caption{Accuracy against artifact size for the fine-tuning adapters.}
  \label{fig:finetune-baselines}
\end{figure}

\paragraph{Comparison to LoRA and VeRA.} We compare our approach to parameter-efficient fine-tuning methods like LoRA~\cite{hu2022lora} and VeRA~\cite{kopiczko2024vera}. Both baselines adapt the same weight matrices and are evaluated across different ranks using PTQ, while our adapter uses QAT. Results are in Table~\ref{tab:finetune-baselines} and Figure~\ref{fig:finetune-baselines}.

\begin{table}[t]
  \caption{Adapter comparison on ResNet-50 $\to$ Oxford-IIIT Pets: trainable parameters and accuracy with artifact size at full precision and at four bits.}
  \label{tab:finetune-baselines}
  \centering
  \small
  \setlength{\tabcolsep}{3pt}
  \begin{tabular}{l r r r r r}
    \toprule
    Method & \#\,train & fp32 & fp32 & int4 & int4 \\
     & & acc & size & acc & size \\
    \midrule
    Full fine-tune          & $23.6$M & $91.36$ & $94.3$\,MB & $2.94$ & $12.2$\,MB \\
    \midrule
    LoRA ($r{=}1$)          & $82$k   & $92.31$ & $319$\,KB   & $90.76$ & $40.3$\,KB \\
    LoRA ($r{=}4$)          & $326$k  & $92.78$ & $1.3$\,MB   & $92.53$ & $159.7$\,KB \\
    LoRA ($r{=}16$)         & $1.30$M & $92.86$ & $5.2$\,MB   & $92.67$ & $637.4$\,KB \\
    \midrule
    VeRA ($r{=}16$)         & $27$k   & $89.56$ & $107.9$\,KB & $87.63$ & $13.9$\,KB \\
    VeRA ($r{=}256$)        & $40$k   & $91.50$ & $158.5$\,KB & $90.71$ & $20.2$\,KB \\
    VeRA ($r{=}1024$)       & $82$k   & $92.42$ & $320.5$\,KB & $91.39$ & $40.4$\,KB \\
    \midrule
    SLVT ($d{=}2048$)  & $2$k  & $74.30$ & $8.0$\,KB  & $69.66$ & $1.0$\,KB \\
    SLVT ($d{=}8192$)  & $8$k  & $88.69$ & $32.0$\,KB & $\mathbf{88.36}$ & $\mathbf{4.0}$\,\textbf{KB} \\
    SLVT ($d{=}16384$) & $16$k & $88.93$ & $64.0$\,KB & $87.57$ & $8.0$\,KB \\
    \bottomrule
  \end{tabular}
\end{table}

Our adapter, at four bits, reaches 88.4\% accuracy using just 4 KB, outperforming the smallest VeRA model, which needs 13.9 KB for 87.6\% accuracy. While LoRA and VeRA achieve higher accuracy given larger storage budgets (up to 92.7\% at 0.6 MB), our single latent plateaus near 89\% due to capacity limits. However, our low-dimensional latent survives quantization much better than full weights. For instance, a four-bit adapter maintains nearly 88\% accuracy, whereas a four-bit full network collapses to 2.9\%. Ultimately, our mapped adapter is ideal for strict kilobyte limits, offering a 5 to 160 times smaller artifact at a slight accuracy cost, while LoRA and VeRA are better when larger sizes are acceptable.

\section{Conclusion}
\label{sec:conclusion}

We introduced Kilobyte models, an extreme compression paradigm that stores neural networks as a deterministic seed and a quantized latent vector rather than as a collection of weights. By leveraging a reproducible, block-wise random basis, we successfully decouple the deployable artifact size from the target network's actual parameter count. Our experiments demonstrate that this formulation, particularly when paired with quantization-aware training, matches the accuracy of aggressively quantized full networks at a fraction of the storage cost. From achieving 98.6\% accuracy on MNIST in just 2 KB to fine-tuning a ResNet-50 adapter in 4 KB, Kilobyte models prove highly effective in the ultra-low storage regime. Furthermore, our approach outperforms existing parameter-efficient methods, such as LoRA and VeRA, when constrained to strict kilobyte budgets. Ultimately, this framework demonstrates that trading explicit parameter storage for latent capacity is a powerful strategy for bandwidth-constrained deployments.

\paragraph{Limitations.}
Kilobyte models drastically reduce storage but do not compress inference compute, as regenerating dense weights leaves inference latency and memory unchanged. Second, accuracy is bottlenecked by latent capacity, causing slight performance drops on harder datasets like FashionMNIST and Covertype. Third, fine-tuning adapters requires the prerequisite base model to be locally present on the deployment device. Finally, Mapping Networks are not yet extended to Transformers and Large Language Models, meaning Kilobyte models cannot currently support them. We are actively working on extending both frameworks to these architectures.

\bibliography{aaai2027}


\mbox{} 
\newpage 

\appendix

\section{Appendix: Kilobyte models pseudocode}
\begin{algorithm}[h!]
\caption{Kilobyte model: encoding (training and quantization)}
\label{alg:encode}
\begin{algorithmic}[1]
\Require target architecture $f$, data distribution $\mathcal{D}$, task loss $\mathcal{L}$,
         layer partition $L$ with latent budgets $\{d_{\ell}\}_{\ell=1}^{L}$,
         bit width $b$, activation $\sigma$, modulation coefficient $\alpha$, integer seed $s$
\Ensure  artifact $\mathcal{A}$
\State partition the trainable parameters of $f$ into mapped
       $\theta_{\mathrm{map}} = (\theta^{1},\dots,\theta^{L})$, with $\theta^{\ell} \in \R^{P_{\ell}}$,
       and unmapped $\theta_{\mathrm{norm}} \in \R^{P_{n}}$
       \Comment{$L=1$ is SLVT; $L>1$ is LWT}
\For{$\ell = 1$ \textbf{to} $L$}
  \State $(W_{0}^{\ell},\, b_{0}^{\ell}) \gets \rho(s, \ell)$
         \Comment{generated from the seed, never stored}
\EndFor
\State $\vz_{\ell} \gets \epsilon\,\xi_{\ell}$ with $\xi_{\ell}\sim\mathcal{N}(0,I_{d_{\ell}})$ for all $\ell$;
       \; $\theta_{\mathrm{norm}} \gets$ standard initialization
\Statex \textit{Latent training: only $\vz$ and $\theta_{\mathrm{norm}}$ receive gradients.}
\Repeat
  \State draw a minibatch $(x,y) \sim \mathcal{D}$
  \For{$\ell = 1$ \textbf{to} $L$}
    \State $\theta^{\ell} \gets \sigma\!\left(W_{0}^{\ell}\vz_{\ell}
            + \alpha \norm{\vz_{\ell}}_{2}^{2}\vone + b_{0}^{\ell}\right)$
  \EndFor
  \State $J \gets \mathcal{L}\!\left(f_{(\theta_{\mathrm{map}},\,\theta_{\mathrm{norm}})}(x),\, y\right)$
  \State update $\{\vz_{\ell}\}$ and $\theta_{\mathrm{norm}}$ by gradient descent on $J$
         \Comment{$W_{0}^{\ell}, b_{0}^{\ell}, \alpha$ frozen}
\Until{converged, giving $\{\vz_{\ell}^{\star}\}$}
\Statex \textit{Quantization to $b$ bits, one scale per layer.}
\If{post-training quantization}
  \State $\hat{\vz}_{\ell} \gets Q_{b}(\vz_{\ell}^{\star})$ for all $\ell$
\Else \Comment{quantization-aware training}
  \Repeat
    \State draw $(x,y) \sim \mathcal{D}$; \;
           $\tilde{\vz}_{\ell} \gets \vz_{\ell} + \bigl(Q_{b}(\vz_{\ell}) - \vz_{\ell}\bigr)_{\mathrm{detach}}$
           \Comment{straight-through estimator}
    \State evaluate $J$ as above using $\tilde{\vz}_{\ell}$ and update $\{\vz_{\ell}\}$
  \Until{converged}
  \State $\hat{\vz}_{\ell} \gets Q_{b}(\vz_{\ell})$ for all $\ell$
\EndIf
\State \Return $\mathcal{A} = \bigl(s,\; \{\hat{\vz}_{\ell}\}_{\ell=1}^{L},\; \theta_{\mathrm{norm}}\bigr)$,
       of size $|\mathcal{A}| = Db/8 + 4L + 8 + 4P_{n}$ bytes, where $D = \sum_{\ell} d_{\ell}$
\end{algorithmic}
\end{algorithm}

\begin{algorithm}[t]
\caption{Kilobyte model: decoding (deployment on the receiving device)}
\label{alg:decode}
\begin{algorithmic}[1]
\Require artifact $\mathcal{A} = \bigl(s, \{\hat{\vz}_{\ell}\}_{\ell=1}^{L}, \theta_{\mathrm{norm}}\bigr)$,
         target architecture $f$, activation $\sigma$, modulation coefficient $\alpha$
\Ensure  a ready-to-run network $f_{\theta}$
\For{$\ell = 1$ \textbf{to} $L$}
  \State $(W_{0}^{\ell},\, b_{0}^{\ell}) \gets \rho(s, \ell)$
         \Comment{same seed reproduces the same objects}
  \State $\vz_{\ell} \gets \Delta_{\ell} \cdot \mathrm{int}(\hat{\vz}_{\ell})$
         \Comment{dequantize with the stored per-layer scale}
  \State $\theta^{\ell} \gets \sigma\!\left(W_{0}^{\ell}\vz_{\ell}
          + \alpha \norm{\vz_{\ell}}_{2}^{2}\vone + b_{0}^{\ell}\right)$
\EndFor
\State $\theta \gets \bigl((\theta^{1},\dots,\theta^{L}),\, \theta_{\mathrm{norm}}\bigr)$
\State \Return $f_{\theta}$
       \Comment{inference is an ordinary dense forward pass}
\end{algorithmic}
\end{algorithm}

\paragraph{Remark (regimes).} $L=1$ recovers single latent vector training, where one
latent generates every mapped parameter, and $L>1$ recovers layer-wise training with
one latent per layer. Only the loop bounds change.

\paragraph{Remark (fine-tuning).} To adapt a pretrained network, replace the
seed-generated center $b_{0}^{\ell}$ by the corresponding block of the pretrained
parameters $\theta_{\mathrm{pre}}$ and take $\sigma$ to be the identity, so that the
latent encodes a displacement $\theta_{\mathrm{ft}} = \theta_{\mathrm{pre}} + W_{0}\vz$.
The artifact then carries a short identifier $h$ of the base model,
$\mathcal{A}_{\mathrm{ft}} = (h, s, \{\hat{\vz}_{\ell}\}, \theta_{\mathrm{norm}})$, and the
base is referenced rather than stored.

\section{Appendix: Hyperparameter values for the experiments for reproducibility}

\paragraph{CNN.} Latents are trained for $30$ epochs with Adam at learning rate
$10^{-2}$ and batch size $128$, under a cosine schedule with $100$ warmup steps
decaying to $5\%$ of the peak rate, followed by $2$ epochs of
quantization-aware fine-tuning with Adam at $2\times10^{-3}$ under a cosine
schedule without warmup; the full-network baselines use Adam at $10^{-3}$, batch size $256$, and $20$ epochs with $200$ warmup
steps. All forward and backward passes run under \texttt{bfloat16} autocast, and no
gradient clipping is used.

\paragraph{LSTM.} Latents are trained for $60$ epochs with Adam at learning rate
$10^{-2}$ and batch size $256$, under a cosine schedule with $100$ warmup steps
decaying to $5\%$ of the peak rate, followed by $3$ epochs of
quantization-aware fine-tuning with Adam at $2\times10^{-3}$ at a constant rate;
the full-network baseline uses Adam at $10^{-3}$, batch size
$256$, and $40$ epochs with $100$ warmup steps. All forward and backward passes run
under \texttt{bfloat16} autocast, and no gradient clipping is used.

\paragraph{MLP.} Latents are trained for $30$ epochs with Adam at learning rate
$10^{-2}$ and batch size $512$, under a cosine schedule with $100$ warmup steps
decaying to $5\%$ of the peak rate, followed by $2$ epochs of
quantization-aware fine-tuning with Adam at $2\times10^{-3}$ under a cosine
schedule without warmup; the full-network baselines use AdamW at $10^{-3}$, batch size $512$, and $20$ epochs with $200$ warmup
steps. All forward and backward passes run under \texttt{bfloat16} autocast, and no
gradient clipping is used.

\paragraph{Fine-tuning.} Latents are trained for $30$ epochs with Adam at learning
rate $5\times10^{-2}$ and batch size $32$, under a cosine schedule with $100$ warmup
steps decaying to $5\%$ of the peak rate, with gradients clipped
to norm $1.0$, which is required for stability at the largest latent budget,
followed by $2$ epochs of quantization-aware fine-tuning with Adam at
$5\times10^{-3}$ at a constant rate and the same clipping; the full fine-tuning
baseline uses Adam at $3\times10^{-4}$, batch size $32$, and
$15$ epochs with $100$ warmup steps. All forward and backward passes run under
\texttt{bfloat16} autocast.

\section{Appendix: Experiment results with 3 seeds variability}

The tables in this Section repeat the results of Section 5 with a standard deviation attached to every reported quantity. The reported value in each case is the same mean given in the main text, with the standard deviation reported across 3 random seeds. A random seed redraws the trained latent, the batch order, and the seeded basis and center. Artifact sizes are deterministic and carry no variability, so they are not repeated here.

\begin{table*}[t]
\caption{Near-orthonormality of the seeded basis: Gram statistics of $W_{0}^{\top}W_{0}$ over a sample of $256$ columns, for the Gaussian and Rademacher constructions at three parameter counts $P$.}
  \label{tab:basis-std}
  \centering
  \small
  \setlength{\tabcolsep}{5pt}
  \begin{tabular}{lcccc}
    \toprule
    Construction & P & Mean col- & Mean           & Max            \\
                 &  & umn norm & $|$off-diag$|$ & $|$off-diag$|$ \\
    \midrule
    Gaussian &$105{,}866$ & $0.99983 \pm 0.000093$ & $0.00245 \pm 0.000002$ & $0.01242 \pm 0.000244$ \\
    Gaussian &$538{,}081$ & $0.99991 \pm 0.000071$ & $0.00108 \pm 0.000006$ & $0.00571 \pm 0.000282$ \\
    Gaussian &$5{,}044{,}942$ & $0.99998 \pm 0.000012$ & $0.00035 \pm 0.000001$ & $0.00177 \pm 0.000052$ \\
    \midrule
    Rademacher &$105{,}866$ & $0.99999 \pm 0.000000$ & $0.00244 \pm 0.000004$ & $0.01319 \pm 0.000166$ \\
    Rademacher &$538{,}081$ & $1.00010 \pm 0.000000$ & $0.00109 \pm 0.000004$ & $0.00595 \pm 0.000177$ \\
    Rademacher &$5{,}044{,}942$ & $0.99905 \pm 0.000000$ & $0.00035 \pm 0.000000$ & $0.00180 \pm 0.000027$ \\
    \bottomrule
  \end{tabular}
\end{table*}

\begin{table*}[t]
\caption{CNN Results: Test accuracy (mean\% $\pm$ standard deviation\%) for the two targets trained conventionally (``full'') and as mapped Kilobyte models under SLVT and LWT, at latent budget $d$ and int8/int4 under PTQ and QAT, on MNIST and FashionMNIST.}
  \label{tab:cnn-std}
  \centering
  \small
  \setlength{\tabcolsep}{3pt}
  \begin{tabular}{l r c c c c c}
    \toprule
    & & \multicolumn{5}{c}{Test accuracy (\%)} \\
    \cmidrule(lr){3-7}
    & & fp32 & \multicolumn{2}{c}{int8} & \multicolumn{2}{c}{int4} \\
    \cmidrule(lr){4-5}\cmidrule(lr){6-7}
    Model & $d$ & & PTQ & QAT & PTQ & QAT \\
    \midrule
    \multicolumn{7}{l}{\emph{MNIST}} \\
    \shortstack[l]{CNN1 full \\ $P=538{,}081$} & --- & $99.20 \pm 0.01$ & $99.18 \pm 0.01$ & --- & $99.12 \pm 0.01$ & --- \\
    CNN1 SLVT & $1024$ & $95.97 \pm 0.09$ & $95.85 \pm 0.11$ & $96.00 \pm 0.09$ & $87.12 \pm 2.99$ & $95.56 \pm 0.26$ \\
    CNN1 SLVT & $2048$ & $97.73 \pm 0.01$ & $97.74 \pm 0.02$ & $97.80 \pm 0.00$ & $87.16 \pm 4.39$ & $97.57 \pm 0.11$ \\
    CNN1 SLVT & $4096$ & $98.51 \pm 0.01$ & $98.57 \pm 0.03$ & $98.61 \pm 0.06$ & $96.20 \pm 0.33$ & $98.45 \pm 0.03$ \\
    CNN1 LWT  & $1024$ & $96.09 \pm 0.08$ & $96.11 \pm 0.08$ & $96.00 \pm 0.20$ & $92.41 \pm 0.21$ & $95.61 \pm 0.12$ \\
    CNN1 LWT  & $2048$ & $97.56 \pm 0.12$ & $97.56 \pm 0.10$ & $97.53 \pm 0.16$ & $95.65 \pm 0.23$ & $97.26 \pm 0.14$ \\
    CNN1 LWT  & $4096$ & $98.06 \pm 0.13$ & $98.07 \pm 0.13$ & $98.08 \pm 0.16$ & $97.09 \pm 1.48$ & $97.94 \pm 0.11$ \\
    \shortstack[l]{CNN2 full \\ $P=105{,}866$} & --- & $98.99 \pm 0.02$ & $98.99 \pm 0.02$ & --- & $98.72 \pm 0.02$ & --- \\
    CNN2 SLVT & $1024$ & $96.73 \pm 0.00$ & $96.73 \pm 0.01$ & $96.68 \pm 0.02$ & $91.09 \pm 0.26$ & $96.41 \pm 0.01$ \\
    CNN2 SLVT & $2048$ & $98.10 \pm 0.07$ & $98.13 \pm 0.10$ & $98.08 \pm 0.03$ & $94.46 \pm 0.73$ & $97.94 \pm 0.14$ \\
    CNN2 SLVT & $4096$ & $98.66 \pm 0.00$ & $98.67 \pm 0.03$ & $98.61 \pm 0.02$ & $97.49 \pm 0.34$ & $\mathbf{98.60} \pm 0.07$ \\
    CNN2 LWT  & $1024$ & $96.49 \pm 0.10$ & $96.38 \pm 0.16$ & $96.37 \pm 0.21$ & $94.60 \pm 2.70$ & $95.79 \pm 0.17$ \\
    CNN2 LWT  & $2048$ & $97.81 \pm 0.05$ & $97.80 \pm 0.08$ & $97.72 \pm 0.13$ & $96.05 \pm 1.66$ & $97.46 \pm 0.11$ \\
    CNN2 LWT  & $4096$ & $98.46 \pm 0.03$ & $98.47 \pm 0.03$ & $98.42 \pm 0.02$ & $96.83 \pm 0.27$ & $98.22 \pm 0.06$ \\
    \midrule
    \multicolumn{7}{l}{\emph{FashionMNIST}} \\
    \shortstack[l]{CNN1 full \\ $P=538{,}081$} & --- & $92.45 \pm 0.16$ & $92.46 \pm 0.16$ & --- & $91.47 \pm 1.17$ & --- \\
    CNN1 SLVT & $1024$ & $85.55 \pm 0.45$ & $85.50 \pm 0.40$ & $85.49 \pm 0.42$ & $78.16 \pm 4.18$ & $84.97 \pm 0.48$ \\
    CNN1 SLVT & $2048$ & $86.86 \pm 0.24$ & $86.90 \pm 0.23$ & $86.81 \pm 0.12$ & $78.27 \pm 2.88$ & $86.67 \pm 0.48$ \\
    CNN1 SLVT & $4096$ & $88.70 \pm 0.08$ & $88.62 \pm 0.10$ & $88.83 \pm 0.06$ & $85.05 \pm 0.43$ & $88.16 \pm 0.16$ \\
    CNN1 LWT  & $1024$ & $85.34 \pm 0.10$ & $85.21 \pm 0.16$ & $85.29 \pm 0.06$ & $78.26 \pm 0.47$ & $84.51 \pm 0.01$ \\
    CNN1 LWT  & $2048$ & $86.76 \pm 0.31$ & $86.55 \pm 0.48$ & $86.97 \pm 0.24$ & $79.31 \pm 1.64$ & $86.49 \pm 0.14$ \\
    CNN1 LWT  & $4096$ & $88.13 \pm 0.12$ & $88.08 \pm 0.17$ & $88.27 \pm 0.06$ & $81.36 \pm 2.28$ & $87.90 \pm 0.25$ \\
    \shortstack[l]{CNN2 full \\ $P=105{,}866$} & --- & $91.06 \pm 0.15$ & $91.02 \pm 0.11$ & --- & $89.28 \pm 2.55$ & --- \\
    CNN2 SLVT & $1024$ & $85.88 \pm 0.30$ & $85.81 \pm 0.30$ & $85.96 \pm 0.37$ & $74.16 \pm 2.51$ & $85.40 \pm 0.30$ \\
    CNN2 SLVT & $2048$ & $87.80 \pm 0.23$ & $87.85 \pm 0.23$ & $87.84 \pm 0.27$ & $85.13 \pm 3.59$ & $87.74 \pm 0.24$ \\
    CNN2 SLVT & $4096$ & $89.52 \pm 0.31$ & $89.46 \pm 0.26$ & $89.46 \pm 0.12$ & $87.84 \pm 2.15$ & $\mathbf{89.11} \pm 0.12$ \\
    CNN2 LWT  & $1024$ & $85.88 \pm 0.01$ & $85.74 \pm 0.08$ & $85.89 \pm 0.10$ & $82.67 \pm 1.26$ & $85.06 \pm 0.15$ \\
    CNN2 LWT  & $2048$ & $87.72 \pm 0.09$ & $87.75 \pm 0.10$ & $87.92 \pm 0.19$ & $83.75 \pm 1.61$ & $87.19 \pm 0.12$ \\
    CNN2 LWT  & $4096$ & $88.67 \pm 0.12$ & $88.61 \pm 0.02$ & $88.81 \pm 0.02$ & $87.15 \pm 1.05$ & $88.32 \pm 0.12$ \\
    \bottomrule
  \end{tabular}
\end{table*}

\begin{table*}[t]
\caption{LSTM Results on the Beijing air-quality dataset: test MSE ($\times10^{-3}$) for the LSTM trained conventionally (``full'') and as a mapped Kilobyte model under SLVT and LWT, at latent budget $d$ and int8/int4 under PTQ and QAT.}
  \label{tab:lstm-std}
  \centering
  \small
  \setlength{\tabcolsep}{3pt}
  \begin{tabular}{l r c c c c c}
    \toprule
    & & \multicolumn{5}{c}{Test MSE (lower is better)} \\
    \cmidrule(lr){3-7}
    & & fp32 & \multicolumn{2}{c}{int8} & \multicolumn{2}{c}{int4} \\
    \cmidrule(lr){4-5}\cmidrule(lr){6-7}
    Model & $d$ & & PTQ & QAT & PTQ & QAT \\
    \midrule
    \shortstack[l]{LSTM full \\ $P=12{,}949$} & --- & $0.00056 \pm 0.000005$ & $0.00056 \pm 0.000005$ & --- & $0.00076 \pm 0.000145$ & --- \\
    LSTM SLVT & $64$   & $0.00071 \pm 0.000040$ & $0.00071 \pm 0.000045$ & $0.00071 \pm 0.000045$ & $0.00090 \pm 0.002640$ & $0.00085 \pm 0.000000$ \\
    LSTM SLVT & $256$  & $0.00060 \pm 0.000005$ & $0.00060 \pm 0.000005$ & $0.00060 \pm 0.000005$ & $0.00154 \pm 0.003470$ & $0.00061 \pm 0.000015$ \\
    LSTM SLVT & $1024$ & $0.00058 \pm 0.000000$ & $0.00058 \pm 0.000000$ & $0.00059 \pm 0.000005$ & $0.00126 \pm 0.000155$ & $0.00064 \pm 0.000005$ \\
    LSTM SLVT & $2048$ & $0.00059 \pm 0.000010$ & $0.00058 \pm 0.000005$ & $0.00062 \pm 0.000015$ & $0.00081 \pm 0.000060$ & $0.00061 \pm 0.000005$ \\
    LSTM LWT  & $64$   & $0.00078 \pm 0.000000$ & $0.00078 \pm 0.000000$ & $0.00080 \pm 0.000025$ & $0.00232 \pm 0.000515$ & $0.00858 \pm 0.003470$ \\
    LSTM LWT  & $256$  & $0.00059 \pm 0.000015$ & $0.00061 \pm 0.000010$ & $0.00060 \pm 0.000005$ & $0.01601 \pm 0.005650$ & $0.00063 \pm 0.000015$ \\
    LSTM LWT  & $1024$ & $0.00056 \pm 0.000005$ & $0.00057 \pm 0.000000$ & $0.00059 \pm 0.000010$ & $0.00064 \pm 0.000830$ & $0.00059 \pm 0.000020$ \\
    LSTM LWT  & $2048$ & $0.00055 \pm 0.000010$ & $0.00056 \pm 0.000005$ & $0.00058 \pm 0.000010$ & $0.00062 \pm 0.000105$ & $\mathbf{0.00058} \pm 0.000035$ \\
    \bottomrule
  \end{tabular}
\end{table*}

\begin{table*}[t]
\caption{MLP Results on two tabular datasets: test accuracy (mean\% $\pm$ standard deviation\%) for MLP1 and MLP2 trained conventionally (``full'') and as mapped Kilobyte models under SLVT and LWT, at latent budget $d$ and int8/int4 under PTQ and QAT.}
  \label{tab:mlp-std}
  \centering
  \small
  \setlength{\tabcolsep}{3pt}
  \begin{tabular}{l r c c c c c}
    \toprule
    & & \multicolumn{5}{c}{Test accuracy (\%)} \\
    \cmidrule(lr){3-7}
    & & fp32 & \multicolumn{2}{c}{int8} & \multicolumn{2}{c}{int4} \\
    \cmidrule(lr){4-5}\cmidrule(lr){6-7}
    Model & $d$ & & PTQ & QAT & PTQ & QAT \\
    \midrule
    \multicolumn{7}{l}{\emph{HIGGS} (binary)} \\
    \shortstack[l]{MLP1 full \\ $P=5{,}044{,}942$} & --- & $73.06 \pm 0.02$ & $73.06 \pm 0.00$ & --- & $72.02 \pm 0.05$ & --- \\
    MLP1 SLVT & $4096$ & $70.53 \pm 0.08$ & $70.52 \pm 0.08$ & $70.52 \pm 0.12$ & $69.24 \pm 0.60$ & $70.28 \pm 0.02$ \\
    MLP1 SLVT & $8192$ & $71.36 \pm 0.08$ & $71.36 \pm 0.09$ & $71.38 \pm 0.05$ & $68.86 \pm 0.80$ & $71.10 \pm 0.10$ \\
    MLP1 SLVT & $16384$ & $72.34 \pm 0.01$ & $72.34 \pm 0.00$ & $72.33 \pm 0.01$ & $70.85 \pm 0.02$ & $72.10 \pm 0.03$ \\
    MLP1 LWT  & $4096$ & $70.28 \pm 0.05$ & $70.33 \pm 0.05$ & $70.27 \pm 0.07$ & $68.71 \pm 0.26$ & $70.01 \pm 0.00$ \\
    MLP1 LWT  & $8192$ & $71.13 \pm 0.05$ & $71.14 \pm 0.03$ & $71.17 \pm 0.01$ & $70.51 \pm 0.09$ & $71.08 \pm 0.04$ \\
    MLP1 LWT  & $16384$ & $71.82 \pm 0.06$ & $71.83 \pm 0.07$ & $71.90 \pm 0.02$ & $69.77 \pm 0.28$ & $71.68 \pm 0.12$ \\
    \shortstack[l]{MLP2 full \\ $P=1{,}003{,}102$} & --- & $73.87 \pm 0.00$ & $73.84 \pm 0.03$ & --- & $72.14 \pm 0.32$ & --- \\
    MLP2 SLVT & $4096$ & $71.24 \pm 0.05$ & $71.20 \pm 0.06$ & $71.19 \pm 0.06$ & $70.21 \pm 0.06$ & $70.94 \pm 0.04$ \\
    MLP2 SLVT & $8192$ & $72.26 \pm 0.06$ & $72.22 \pm 0.08$ & $72.24 \pm 0.07$ & $71.14 \pm 0.52$ & $72.07 \pm 0.01$ \\
    MLP2 SLVT & $16384$ & $73.01 \pm 0.00$ & $72.99 \pm 0.03$ & $72.94 \pm 0.04$ & $70.45 \pm 0.34$ & $\mathbf{72.73} \pm 0.11$ \\
    MLP2 LWT  & $4096$ & $70.95 \pm 0.15$ & $70.92 \pm 0.12$ & $70.95 \pm 0.16$ & $70.28 \pm 0.41$ & $70.74 \pm 0.11$ \\
    MLP2 LWT  & $8192$ & $71.70 \pm 0.02$ & $71.72 \pm 0.02$ & $71.70 \pm 0.01$ & $70.70 \pm 0.00$ & $71.55 \pm 0.03$ \\
    MLP2 LWT  & $16384$ & $72.23 \pm 0.08$ & $72.25 \pm 0.07$ & $72.29 \pm 0.07$ & $70.98 \pm 0.16$ & $72.05 \pm 0.13$ \\
    \midrule
    \multicolumn{7}{l}{\emph{Covertype} (7-class)} \\
    \shortstack[l]{MLP1 full \\ $P=5{,}093{,}927$} & --- & $96.65 \pm 0.01$ & $96.62 \pm 0.01$ & --- & $88.16 \pm 0.46$ & --- \\
    MLP1 SLVT & $4096$ & $82.12 \pm 0.12$ & $81.88 \pm 0.21$ & $82.10 \pm 0.15$ & $70.42 \pm 0.85$ & $79.43 \pm 0.13$ \\
    MLP1 SLVT & $8192$ & $84.71 \pm 0.08$ & $84.60 \pm 0.09$ & $84.82 \pm 0.03$ & $69.49 \pm 0.68$ & $82.02 \pm 0.13$ \\
    MLP1 SLVT & $16384$ & $87.76 \pm 0.01$ & $87.64 \pm 0.05$ & $87.94 \pm 0.06$ & $75.96 \pm 0.12$ & $85.25 \pm 0.06$ \\
    MLP1 LWT  & $4096$ & $82.68 \pm 0.03$ & $82.50 \pm 0.02$ & $82.71 \pm 0.05$ & $73.44 \pm 0.38$ & $79.99 \pm 0.02$ \\
    MLP1 LWT  & $8192$ & $84.85 \pm 0.05$ & $84.80 \pm 0.02$ & $84.90 \pm 0.06$ & $66.99 \pm 3.48$ & $82.37 \pm 0.30$ \\
    MLP1 LWT  & $16384$ & $87.31 \pm 0.04$ & $87.28 \pm 0.04$ & $87.39 \pm 0.07$ & $74.76 \pm 0.77$ & $84.73 \pm 0.26$ \\
    \shortstack[l]{MLP2 full \\ $P=1{,}024{,}807$} & --- & $96.13 \pm 0.01$ & $96.12 \pm 0.01$ & --- & $83.43 \pm 0.30$ & --- \\
    MLP2 SLVT & $4096$ & $83.43 \pm 0.02$ & $83.36 \pm 0.02$ & $83.53 \pm 0.04$ & $70.50 \pm 0.24$ & $80.13 \pm 0.03$ \\
    MLP2 SLVT & $8192$ & $86.62 \pm 0.22$ & $86.52 \pm 0.21$ & $86.73 \pm 0.20$ & $65.50 \pm 2.62$ & $83.37 \pm 0.06$ \\
    MLP2 SLVT & $16384$ & $91.04 \pm 0.07$ & $90.77 \pm 0.11$ & $91.18 \pm 0.07$ & $73.04 \pm 0.73$ & $\mathbf{86.64} \pm 0.25$ \\
    MLP2 LWT  & $4096$ & $83.29 \pm 0.16$ & $83.03 \pm 0.12$ & $83.35 \pm 0.16$ & $71.42 \pm 2.41$ & $80.28 \pm 0.30$ \\
    MLP2 LWT  & $8192$ & $86.23 \pm 0.07$ & $86.16 \pm 0.09$ & $86.38 \pm 0.12$ & $74.83 \pm 3.21$ & $81.88 \pm 0.27$ \\
    MLP2 LWT  & $16384$ & $89.46 \pm 0.05$ & $89.38 \pm 0.06$ & $89.63 \pm 0.02$ & $72.75 \pm 0.64$ & $85.24 \pm 0.24$ \\
    \bottomrule
  \end{tabular}
\end{table*}

\begin{table*}[t]
\caption{Fine-tuning Results: Pets test accuracy (\%) for an ImageNet-pretrained ResNet-50 adapted by a mapped delta under SLVT and LWT, at latent budget $d$ and int8/int4 under PTQ and QAT. The baseline is full fine-tuning.}
  \label{tab:finetune-std}
  \centering
  \small
  \setlength{\tabcolsep}{3pt}
  \begin{tabular}{l r c c c c c}
    \toprule
    & & \multicolumn{5}{c}{Test accuracy (\%)} \\
    \cmidrule(lr){3-7}
    & & fp32 & \multicolumn{2}{c}{int8} & \multicolumn{2}{c}{int4} \\
    \cmidrule(lr){4-5}\cmidrule(lr){6-7}
    Method & $d$ & & PTQ & QAT & PTQ & QAT \\
    \midrule
    Full fine-tune & --- & $91.36 \pm 0.26$ & $91.20 \pm 0.06$ & --- & $2.94 \pm 0.10$ & --- \\
    \midrule
    SLVT & $2048$  & $74.30 \pm 0.08$ & $74.52 \pm 0.18$ & $74.03 \pm 0.33$ & $62.96 \pm 2.24$ & $69.66 \pm 0.02$ \\
    SLVT & $8192$  & $88.69 \pm 0.34$ & $88.69 \pm 0.23$ & $88.31 \pm 0.53$ & $86.26 \pm 0.37$ & $\mathbf{88.36} \pm 0.40$ \\
    SLVT & $16384$ & $88.93 \pm 0.12$ & $89.02 \pm 0.07$ & $88.72 \pm 1.44$ & $88.14 \pm 0.08$ & $87.57 \pm 0.23$ \\
    LWT  & $2048$  & $70.99 \pm 1.90$ & $70.97 \pm 10.94$ & $70.70 \pm 10.99$ & $67.04 \pm 1.16$ & $67.26 \pm 1.29$ \\
    LWT  & $8192$  & $87.38 \pm 0.43$ & $87.35 \pm 0.48$ & $87.05 \pm 0.84$ & $85.88 \pm 0.41$ & $86.78 \pm 0.44$ \\
    LWT  & $16384$ & $87.54 \pm 0.42$ & $87.38 \pm 0.55$ & $87.46 \pm 0.39$ & $86.05 \pm 0.59$ & $86.32 \pm 0.34$ \\
    \bottomrule
  \end{tabular}
\end{table*}

\end{document}